\documentclass[twoside]{article}

\usepackage{PRIMEarxiv}

\usepackage[utf8]{inputenc} 
\usepackage[T1]{fontenc}    
\usepackage{hyperref}       
\usepackage{url}            
\usepackage{booktabs}       
\usepackage{amsfonts}       
\usepackage{nicefrac}  
\usepackage{float}    
\usepackage{tikz}
\usetikzlibrary{shapes.geometric, arrows}

\tikzstyle{startstop} = [rectangle, rounded corners, minimum width=0.5cm, minimum height=1cm,text centered, draw=black, fill=violet!40]
\tikzstyle{process} = [rectangle, minimum width=1cm, minimum height=0.5cm, text centered, draw=black, fill=violet!30]
\tikzstyle{decision} = [rectangle, minimum width=1cm, minimum height=0.5cm, text centered, draw=black, fill=red!40]
\tikzstyle{data} = [rectangle, minimum width=1cm, minimum height=0.5cm, text centered, draw=black, fill=green!40]
\tikzstyle{result} = [rectangle, minimum width=1cm, minimum height=0.5cm, text centered, draw=black, fill=yellow!40]
\tikzstyle{arrow} = [thick,->,>=stealth]
\usepackage{multirow}
\usepackage{multicol}
\usepackage{array}
\usepackage{microtype}      
\usepackage{lipsum}
\usepackage{fancyhdr}       
\usepackage{graphicx}       
\graphicspath{{media/}}     

\title{Development and Comparative Evaluation of Three Artificial Intelligence Models (NLP, LLM, JEPA) for Predicting Triage in Emergency Departments: A 7-Month Retrospective Proof-of-Concept
}

\author{
  Edouard Lansiaux \\
  Department of Emergency \\
  Lille University Hospital\\
  \texttt{edouard.lansiaux@orange.fr} \\
   \And
  Ramy Azzouz \\
  Centre Antipoison, Lille University Hospital\\
  ULR 2694-METRICS, Lille University\\
  \texttt{ramy.azzouz@chu-lille.fr} \\
  \AND
   Emmanuel Chazard \\
   Department of Public Health, EA 2694 \\
   \& ULR 2694-METRICS\\
   Lille University\\
  \And
  Amélie Vromant \\
  Emergency Department\\ 
  Hôpital Pitié-Salpêtrière, AP-HP ,\\
  Paris, France\\
  \And
  Eric Wiel \\
    Department of Emergency , Lille University Hospital \\
  ULR 2694-METRICS, Lille University\\
}

\begin{document}
\maketitle

\begin{abstract}
Triage errors, including undertriage and overtriage, are persistent challenges in emergency departments (EDs). With increasing patient influx and staff shortages, the integration of artificial intelligence (AI) into triage protocols has gained attention. This study compares the performance of three AI models: Natural Language Processing (NLP), Large Language Models (LLM), and Joint Embedding Predictive Architecture (JEPA) to predict triage outcomes against the FRENCH scale and clinical practice. We conducted a retrospective analysis of a prospectively recruited cohort based on adult patient triage data over a 7-month period at Roger Salengro Hospital ED (Lille, France). Three AI models were trained and validated: (1) TRIAGEMASTER (NLP), (2) URGENTIAPARSE (LLM) and (3) EMERGINET (JEPA).  Data included demographic details, verbatim chief complaints, vital signs, and triage outcomes based on the FRENCH scale and GEMSA coding. The primary outcome was the concordance of the AI-predicted triage level with the French gold standard. It was assessed thanks to various indicators: F1-Score, Weighted Kappa, Spearman, MAE, RMSE, AUC-ROC. The LLM model (URGENTIAPARSE) showed higher accuracy (composite score: 2.514) compared to JEPA (EMERGINET, 0.438) and NLP (TRIAGEMASTER, -3.511), outperforming nurse triage (-4.343). This observation is reinforced by the F1-Score and AUC-ROC : 0.900 and 0.879 for  URGENTIAPARSE; 0.731 and 0.686 for EMERGINET;  0.618 and 0.642  for TRIAGEMASTER; respectively 0.303 and 0.776 for nurse triage. Secondary analyses highlighted the effectiveness of URGENTIAPARSE in predicting hospitalization needs (GEMSA) and its robustness with structured data versus raw transcripts (either for GEMSA prediction or for FRENCH prediction). LLM architecture, through abstraction of patient representations, offers the most accurate triage predictions among tested models. Integrating AI into ED workflows could enhance patient safety and operational efficiency, though integration into clinical workflows requires addressing model limitations and ensuring ethical transparency.

\end{abstract}

\keywords{Triage \and Emergency Department \and Artificial Intelligence \and Natural Language Processing \and Large Language Model \and Joint Embedding Predictive Architecture.}

\section{Introduction} Triage in emergency departments is vital for prioritizing patients by severity, directly impacting outcomes, efficiency, and resource use. In busy or understaffed settings, accurate triage is crucial for safe and equitable care.

In France, the FRENCH triage scale has become a standardized reference tool for emergency medical services\cite{Taboulet2009TriageWT}. This six-level urgency stratification system was designed to promote consistency and reproducibility. Yet, in practice, its application often varies \cite{Husted2000MethodsFA} —shaped by disparities in nurse experience, time pressure, interpretive inconsistencies, and fatigue. These factors lead to two key errors: undertriage, where serious cases are underestimated, risking delays in care; and overtriage, where minor cases are over-prioritized, causing crowding and inefficient use of resources \cite{Moll2010ChallengesIT}. 

Against this backdrop, Artificial Intelligence (AI) has emerged as a promising complement to traditional triage \cite{Levin2017MachineLearningBasedET}. AI systems in emergency medicine can process diverse data, work tirelessly, and provide consistent, transparent decisions, but their success over the past decade has been mixed. \\
Early AI triage systems relied on Natural Language Processing (NLP) to extract clinical meaning from text, using hand-crafted rules or statistical models \cite{Ilicki2022ChallengesIE}. Although earlier systems worked, they could not understand context, nuance, or rare cases. Advances came with Large Language Models (LLMs) like BERT and its French version, FlauBERT \cite{le-etal-2020-flaubert-unsupervised}. These transformer-based models offered a deep contextual understanding. Combined with boosting algorithms such as XGBoost, LLMs formed hybrid architectures balancing high performance with interpretability \cite{Brossard2025PredictingED}, two pillars for clinical deployment. More recently, Joint Embedding Predictive Architectures (JEPA) have introduced a third paradigm. JEPA models do not predict outcomes directly; they embed both inputs and targets (e.g., triage categories) into a shared latent space8. Using techniques such as energy minimization and variance-invariance-covariance regularization (e.g., VICReg)\cite{bardes2022vicregvarianceinvariancecovarianceregularizationselfsupervised,lansiaux2024navigating}, these models can uncover subtle cross-modal patterns. However, their complexity poses ongoing challenges to interpretability, an essential requirement in medical contexts. \\
Few studies compare AI models under consistent conditions. Most target narrow tasks, use limited or noisy data and are often based outside the French healthcare system, reducing relevance for those using the FRENCH scale\cite{lansiaux2024medical,pmlr-v259-guerra-adames25a}.

The present study aims to address this gap. We conducted a retrospective analysis of triage encounters at the CHU Lille ED, one of France’s major academic hospitals\cite{dress_statistique_nodate}. Using rich clinical data and nurse-patient dialogues, three AI models—TRIAGEMASTER, URGENTIAPARSE, and EMERGINET—were developed to predict FRENCH triage levels. They were evaluated against expert consensus to determine the best performer and assess their clinical suitability and limitations.

\section{Methods}
\subsection{Study Design and Ethical Compliance}
This study was designed as a retrospective, observational, and monocentric analysis carried out in the Adult Emergency Department (ED) of the Center Hospitalier Universitaire (CHU) of Lille, France. The Emergency Department is a high-volume academic tertiary care facility serving a diverse patient population. The investigation covered a continuous seven-month period from June 1, 2024, through December 31, 2024. This time frame was selected to capture seasonal variations in ED activity and to ensure a sufficiently large and heterogeneous data set.

The study followed strict French and European ethical standards, with approval and registration of CESREES, Health Data Hub filing (project number 18605502) and CNIL MR-004 compliance\cite{lansiaux_tiaeu_2023}. No personal data was used, so informed consent was not required. The study followed the TRIPOD-AI guidelines (see Supplementary Data 1). Due to patient information, the full dataset and code are not publicly available: only the code used to define the gold standard is shared.
\subsection{Population and Inclusion Criteria}
The study included adult patients evaluated in the Lille CHU Emergency Department, with complete structured data and recorded nurse-patient audio. Exclusions applied to minors, legally protected adults, and incomplete or corrupted records.

\subsection{Data Acquisition and Preprocessing}

The study relied on two primary data sources. First, structured clinical data were extracted from ResUrgences® ED management software (version 2024.1.148). The variables collected included age, sex, vital signs, the initial decision to select based on the FRENCH scale and the GEMSA scale (corresponding to the patient's outcome after diagnostic/treatment in the emergency room) \cite{dhos_circulaire_2001}.
Unstructured data from nurse-patient triage interviews were manually transcribed, accuracy verified, linked by patient IDs and timestamps, anonymized, and securely stored for modeling.

\subsection{Model Architectures and Training Pipelines}

In this study, three distinct artificial intelligence models were developed and compared, each representing a major class of current machine learning techniques.
\begin{itemize}
\item \textbf{TRIAGEMASTER (NLP)}: This model used paragraph vector embeddings (Doc2Vec) to represent input of free text, concatenated with numerical features derived from structured data \cite{Mikolov2013EfficientEO}. The architecture consisted of a three-layer feedforward neural network with ReLU activations \cite{ramachandran2018searching}, dropout regularization (p=0.05), and L2 norm penalty ($\lambda =1e^{-5}$). Optimization used an initial learning rate of 0.25. The classification level probabilities were derived by testing paragraph vectors, bag-of-words, or topic distributions against models. Confidence intervals for the AUC were calculated with the DeLong method\cite{DeLong1988ComparingTA}. Model performance characteristics were calculated over the full range of predicted sorting probabilities. Optimal probability thresholds were defined as the predicted probability at the maximum F1 score\cite{derczynski-2016-complementarity}.
\item \textbf{URGENTIAPARSE (LLM)}: Built upon the FlauBERT language model \cite{le-etal-2020-flaubert-unsupervised}, this architecture fine-tuned contextual embeddings from patient complaints and combined them with structured variables using XGBoost. The hybrid design allowed both interpretability and high-dimensional pattern recognition \cite{Chrusciel2021ThePO}. Feature importance was derived via SHAP values, providing transparent model behavior traceable to individual tokens or vitals \cite{Lundberg2017AUA}.
\item \textbf{EMERGINET (JEPA)}: This Joint Embedding Predictive Architecture encoded textual and structured inputs into a shared latent space through dual encoders. To process the “history” variable (corresponding to the verbatim of the voice recording corresponding to free text), a neural layer based on FlauBERT was used with a “Long Short Term Memory”\cite{Yu2019ARO}. The loss function minimized contrastive energy between predicted and target classes, regularized using VICReg to enforce statistical stability\cite{bardes2022vicregvarianceinvariancecovarianceregularizationselfsupervised}. 
\end{itemize}
Each model was trained on 80\% of the dataset and validated on the remaining 20\%, stratified by triage level to maintain class balance. Hyperparameter tuning used grid search across predefined ranges.
\subsection{Gold standard Construction}
A gold standard triage label was created by independent reviews from senior clinicians using the full clinical context and FRENCH scale. Disagreements were resolved by consensus, and the final labels served as the benchmark for model training and evaluation.
\subsection{Evaluation Metrics}
Model performance was assessed using both absolute error measures and classification agreement metrics:
\begin{itemize}
\item \textbf{Mean Absolute Error (MAE)} : Measures average deviation between predicted and actual triage levels :
\begin{equation}
MAE = \frac{\sum_{i=1}^{n}|\hat{y}_{i}-y_{i}|}{n}
\end{equation}
where  $\hat{y}_{i}$ is the forecasted value, $y_{i}$ the real value and $n$ the total data number ;
\item \textbf{Root Mean Square Error (RMSE)} : Emphasizes larger misclassification errors :
\begin{equation}
RMSE = \sum_{i=1}^{n}\frac{(\hat{y}_{i}-y_{i})^2}{n}
\end{equation}
where $\hat{y}_{i}$ is the forecasted value, $y_{i}$ the real value and $n$ the total data number ; 
\item \textbf{Weighted Cohen’s Kappa} : Accounts for agreement beyond chance and penalizes larger discrepancies :
\begin{equation}
k_{w}=1-\frac{\sum_{1 \le i,j \le n}{w_{i,j}*fo_{i,j}}}{\sum_{1 \le i,j \le n}w_{i,j}*fe_{i,j}}
\end{equation}
where $fo_{i,j}$ are the observed frequencies, $fe_{i,j}$ the real frequencies, $w$ the ponderation factor and $n$ the total number of data points ;
\item \textbf{Spearman coefficient} : number ranging from -1 to 1 that indicates how strongly two sets of ranks are correlated :
\begin{equation}
r_{s}=\frac{cov(rg_{X},rg_{Y})}{\sigma _{rg_{X}} \sigma _{rg_{Y}}}
\end{equation}
where $cov(rg_{X},rg_{Y})$ is the covariance of rank variables, $\sigma _{rg_{X}}$ and $\sigma _{rg_{Y}}$ are the standard deviations of the rank variables ;
\item \textbf{F1 Scores (micro and macro)} : Macro F1 reflects the unweighted average across all triage classes, treating each class equally, while Micro F1 is the weighted average, accounting for the number of samples per class—together, they show overall precision and recall performance across classes:
\begin{equation}
F_{1}=\frac{TP}{TP + \frac{1}{2}(FP+FN)}
\end{equation}
where TP is the number of true positives, FP is false positives, and FN is false negatives ;
\item \textbf{Percentage of exact agreement} ;
\item \textbf{Percentage of near agreement within ±1 class} ; 
\item \textbf{Area Under Curve - Receiver Operating Curve} is a single number that summarizes the classifier's performance across all possible classification thresholds ;
\item \textbf{Composite Z-score} : an integrated performance indicator combining some previous metrics for ranking:
\begin{equation}
Score = Z (-MAE) + Z(-RMSE) + Z(Weighted \enspace Kappa) + Z (Spearman)
\end{equation}
Where for a metric X:
\begin{equation}
Z(X)=\frac{X-\overline{x}_{X}}{\sigma_{X}}
\end{equation}
where $\overline{x}_{X}$ is the metric mean et $\sigma_{X}$ the standard deviation .
\end{itemize}
Secondary outcomes included Bland-Altman plots for measurement comparison, thermal confusion matrices to analyze class disagreements, error distribution visualizations, and class-based F1 score analysis for each model.
\subsection{Secondary analyses}
A secondary analysis compared actual versus AI-predicted GEMSA scores using models trained on nurse-recorded histories and interview transcripts. Another analysis assessed model performance based on each input type separately. Both analyses used consistent evaluation methods.

We developed an R script to evaluate and visualize the calibration performance of different triage prediction approaches (NLP, LLM, JEPA, and nurse triage [FRENCH inf]) across six ordered triage categories (1 = highest acuity to 6 = lowest acuity). The script first imports the dataset from Excel, detects the gold-standard triage column, and identifies model outputs, either as probability distributions or single predicted labels, which are converted to one-hot encoded probabilities and normalized when required. \\
For each model, a multiclass Brier score \cite{Ahmadian_2025} is calculated as the mean squared difference between predicted probabilities and observed one-hot outcomes, providing a global measure of probabilistic accuracy. \\
Calibration analysis is then performed by grouping predicted probabilities into quantile-based bins for each triage class, comparing mean predicted values with observed frequencies, and producing calibration tables and reliability diagrams against the reference diagonal of perfect calibration. \\ 
To further explore prediction behavior, risk distributions are visualized using density ridge plots stratified by true triage class and faceted by predicted class, allowing assessment of probability allocation across categories. \\
In addition, heatmaps of mean predicted probabilities are generated per model to highlight calibration and misclassification patterns across the six-level triage scale. \\

\section{Results}
\subsection{Population Characteristics}
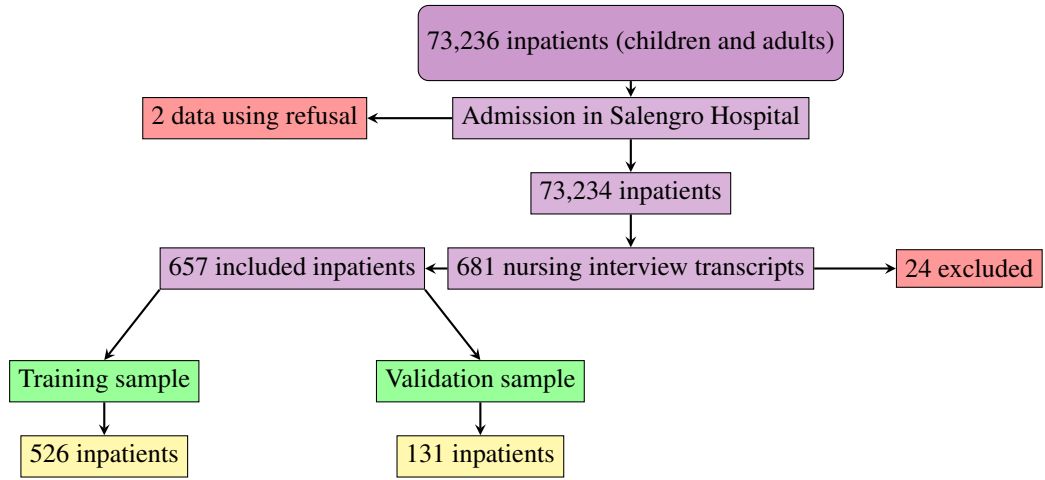
\begin{figure}[!htbp]
    \centering
    \begin{tikzpicture}[node distance=1cm]

    \node (start) [startstop] {73,236 inpatients (children and adults)};

    \node (admission) [process, below of=start] {Admission in Salengro Hospital};

    \node (refusal) [decision, left of=admission, xshift=-4cm] {2 data using refusal};
    \node (inpatients) [process, below of=admission] {73,234 inpatients};

    \node (nursing) [process, below of=inpatients] {681 nursing interview transcripts};

    \node (included) [process, left of=nursing, xshift=-3.5cm] {657 included inpatients};
    \node (excluded) [decision, right of=nursing, xshift=3.5cm] {24 excluded};

    \node (train) [data, below of=included, xshift=-2.5cm, yshift=-0.5cm] {Training sample};
    \node (val) [data, below of=included, xshift=2.5cm, yshift=-0.5cm] {Validation sample};

    \node (trainres) [result, below of=train] {526 inpatients};
    \node (valres) [result, below of=val] {131 inpatients};

    \draw [arrow] (start) -- (admission);
    \draw [arrow] (admission.west) -- (refusal.east);
    \draw [arrow] (admission) -- (inpatients);
    \draw [arrow] (inpatients) -- (nursing);
    \draw [arrow] (nursing.west) -- (included.east);
    \draw [arrow] (nursing.east) -- (excluded.west);
    \draw [arrow] (included.south west) -- (train.north);
    \draw [arrow] (included.south east) -- (val.north);
    \draw [arrow] (train) -- (trainres);
    \draw [arrow] (val) -- (valres);

    \end{tikzpicture}
    \caption{Flowchart of inpatient inclusion, exclusion, and dataset splitting into training and validation samples.}
    \label{fig:fig1}
\end{figure}

During the inclusion period, 73,236 patients visited the emergency department of Roger Salengro Hospital. Two patients declined the use of their data. Triage nurse interviews were recorded and transcribed for 681 patients, representing 0.93\% of the total. After excluding 24 cases, including 6 with missing data, 657 patients (0.897\%) met the inclusion criteria (Figure \ref{fig:fig1}).\\

The mean age of the included patients was 43 years, with a median of 38 (Table ~\ref{tab:table1}). The sample was gender-balanced, with most admissions occurring between 6:00 and 14:00. Main referral reasons were trauma, abdominal, and cardiac issues. A quarter had comorbidities, mostly cardiovascular. Most patients were triaged as sort 3B, with few in higher urgency categories. The dataset was split into training (n=526) and validation (n=131) groups, both with similar characteristics.
\begin{table}[!htbp]
\centering
\begin{tabular}{|l|l|r|r|r|r|r|r|r|}
\hline
\multirow{2}{*}{Population} & \multirow{2}{*}{} & \multicolumn{3}{c|}{} & \multicolumn{2}{c|}{Training} & \multicolumn{2}{c|}{Validation} \\ \cline{3-9}
 &  & Min & Max & Med & N/Mean & \%/±SD & N/Mean & N/Mean \\ \hline
\multirow{3}{*}{Age} &  & 18 & 105 & 38 & 42.6 & 19.71 & 42.5 & 43.0 \\ \cline{2-9}
 & Man & \multicolumn{3}{c|}{} & 330 & 50.23\% & 264 & 66 \\ \cline{2-9}
 & Woman & \multicolumn{3}{c|}{} & 327 & 49.77\% & 262 & 65 \\ \hline
\multirow{3}{*}{Admission hour} & 21:00-06:00 & \multicolumn{3}{c|}{} & 211 & 32.12\% & 169 & 42 \\ \cline{2-9}
 & 06:00-14:00 & \multicolumn{3}{c|}{} & 241 & 36.68\% & 193 & 48 \\ \cline{2-9}
 & 14:00-21:00 & \multicolumn{3}{c|}{} & 205 & 31.20\% & 164 & 41 \\ \hline
\multirow{15}{*}{Medical recourse} & Abdominal & \multicolumn{3}{c|}{} & 102 & 15.53\% & 81 & 21 \\ \cline{2-9}
 & Cardio-circulatory & \multicolumn{3}{c|}{} & 86 & 13.09\% & 68 & 18 \\ \cline{2-9}
 & Various & \multicolumn{3}{c|}{} & 31 & 4.72\% & 24 & 7 \\ \cline{2-9}
 & Genito\-urinary & \multicolumn{3}{c|}{} & 34 & 5.18\% & 27 & 7 \\ \cline{2-9}
 & Gynecology and obstetrics & \multicolumn{3}{c|}{} & 2 & 0.30\% & 1 & 1 \\ \cline{2-9}
 & Infectiology & \multicolumn{3}{c|}{} & 8 & 1.22\% & 6 & 2 \\ \cline{2-9}
 & Poisoning & \multicolumn{3}{c|}{} & 6 & 0.91\% & 4 & 2 \\ \cline{2-9}
 & Neurology & \multicolumn{3}{c|}{} & 69 & 10.50\% & 55 & 14 \\ \cline{2-9}
 & Ophthalmology & \multicolumn{3}{c|}{} & 4 & 0.61\% & 3 & 1 \\ \cline{2-9}
 & ENT/Stomatology & \multicolumn{3}{c|}{} & 62 & 9.44\% & 49 & 13 \\ \cline{2-9}
 & Dermatology & \multicolumn{3}{c|}{} & 40 & 6.08\% & 32 & 8 \\ \cline{2-9}
 & Psychiatry & \multicolumn{3}{c|}{} & 24 & 3.65\% & 19 & 5 \\ \cline{2-9}
 & Pulmonary & \multicolumn{3}{c|}{} & 26 & 3.96\% & 20 & 6 \\ \cline{2-9}
 & Rheumatology & \multicolumn{3}{c|}{} & 36 & 5.48\% & 28 & 8 \\ \cline{2-9}
 & Traumatology & \multicolumn{3}{c|}{} & 127 & 19.33\% & 101 & 26 \\ \hline
\multirow{3}{*}{Comorbidities} & Absence & \multicolumn{3}{c|}{} & 496 & 75.49\% & 398 & 98 \\ \cline{2-9}
 & Presence ($\ge$1) & \multicolumn{3}{c|}{} & 161 & 24.50\% & 128 & 33 \\ \cline{2-9}
 & Vascular ones($\ge$1) & \multicolumn{3}{c|}{} & 75 & 11.42\% & 60 & 15 \\ \hline
\multirow{7}{*}{Vital Signs} & SBP (mmHg) & 84 & 234 & 140 & 143 & 23.67 & 142.7 & 143.3 \\ \cline{2-9}
 & DBP (mmHg) & 0 & 148 & 82 & 83 & 16.15 & 83.1 & 83.2 \\ \cline{2-9}
 & HR (b.p.m.) & 43 & 187 & 87 & 89 & 17.56 & 89.1 & 89 \\ \cline{2-9}
 & Temp. (°C) & 34.9 & 40 & 37.7 & 37.5 & 0.58 & 37.5 & 37.5 \\ \cline{2-9}
 & E.V.A. & 0 & 10 & 3 & 4 & 3.64 & 4.0 & 4.1 \\ \cline{2-9}
 & SpO2 (\%) & 84 & 100 & 97 & 97 & 1.67 & 97.0 & 97.0 \\ \cline{2-9}
 & O2 (L/min) & 0 & 6 & 0 & 0.04 & 0.39 & 0.04 & 0.03 \\ \hline
\multirow{6}{*}{Nurse triage} & 1 & \multicolumn{3}{c|}{} & 4 & 0.61\% & 3 & 1 \\ \cline{2-9}
 & 2 & \multicolumn{3}{c|}{} & 86 & 13.09\% & 68 & 18 \\ \cline{2-9}
 & 3A & \multicolumn{3}{c|}{} & 0 & 0\% & 0 & 0 \\ \cline{2-9}
 & 3B & \multicolumn{3}{c|}{} & 354 & 53.88\% & 283 & 71 \\ \cline{2-9}
 & 4 & \multicolumn{3}{c|}{} & 118 & 17.96\% & 94 & 24 \\ \cline{2-9}
 & 5 & \multicolumn{3}{c|}{} & 95 & 14.46\% & 76 & 18 \\ \hline
\multirow{7}{*}{GEMSA} & 1 & \multicolumn{3}{c|}{} & 0 & 0\% & 0 & 0 \\ \cline{2-9}
 & 2 & \multicolumn{3}{c|}{} & 509 & 77.47\% & 407 & 102 \\ \cline{2-9}
 & 3 & \multicolumn{3}{c|}{} & 18 & 2.74\% & 14 & 4 \\ \cline{2-9}
 & 4 & \multicolumn{3}{c|}{} & 102 & 15.53\% & 82 & 20 \\ \cline{2-9}
 & 5 & \multicolumn{3}{c|}{} & 10 & 1.52\% & 9 & 1 \\ \cline{2-9}
 & 6 & \multicolumn{3}{c|}{} & 0 & 0\% & 0 & 0 \\ \cline{2-9}
 & Not specified & \multicolumn{3}{c|}{} & 18 & 2.74\% & 14 & 4 \\ \hline
\end{tabular}
\caption{Patient demographics, time of admission, reason for referral, comorbidities, vital parameters, nursing triage according to the FRENCH scale, GEMSA scale}
\label{tab:table1}
\end{table}

\subsection{Training Performance}
\begin{figure}[!htbp]
    \centering
    \includegraphics[width=\textwidth]{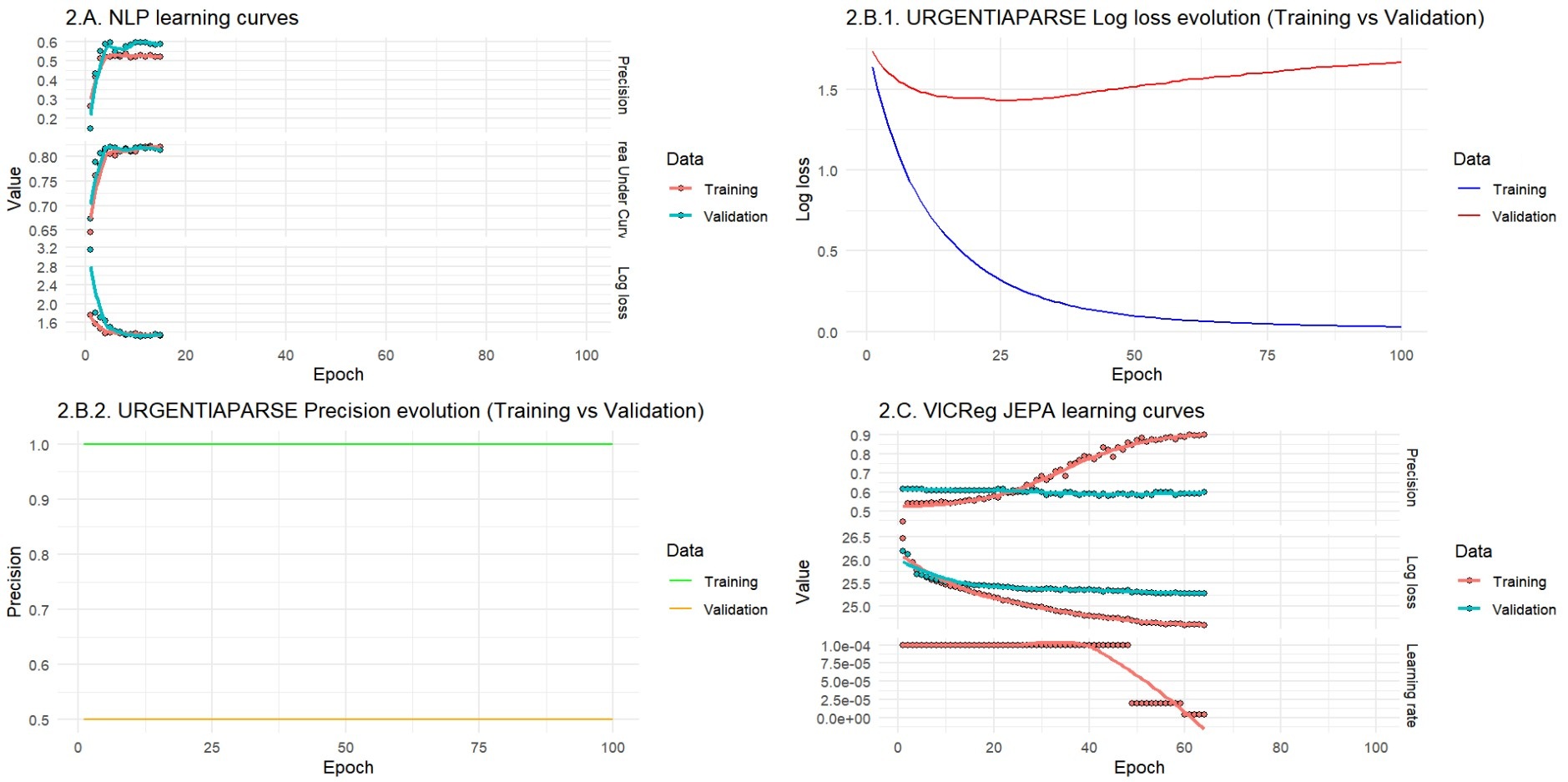}
    \caption{Learning curves for NLP, URGENTIAPARSE, and EMERGINET models. 
    Subfigures show: (A) TRIAGEMASTER learning curves, 
    (B.1) URGENTIAPARSE log loss evolution, 
    (B.2) URGENTIAPARSE precision evolution, 
    and (C) EMERGINET learning curves.}
    \label{fig:fig2}
\end{figure}

Over 100 training iterations, TRIAGEMASTER’s accuracy rose to about 75\%, while log loss decreased and stabilized (Fig. \ref{fig:fig2} ) . URGENTIAPARSE showed perfect training accuracy (1.0) but constant low validation accuracy (0.5), indicating overfitting (Fig. 2.B.2.). Training log loss dropped to near zero, while validation loss remained high and mostly flat and with AUC of 0.6855 (Fig. 2.B.1.). EMERGINET reached \~ 0.85 training accuracy, but validation one stayed at \~ 0.6, indicating poor generalization (Fig. 2.C.). Training loss declined steadily; validation loss remained higher and stable, reflecting performance gaps.

\subsection{Overall Model Performance}
\begin{table}[!htbp]
\centering
\begin{tabular}{l *{10}{c}}
\toprule
Process & MAE & RMSE & Kappa & \begin{tabular}[c]{@{}r@{}} Spear\\ -man \end{tabular} & \begin{tabular}[c]{@{}r@{}} F1\\ Micro\end{tabular} & \begin{tabular}[c]{@{}r@{}} F1\\ Macro\end{tabular} & \begin{tabular}[c]{@{}r@{}r@{}} Exact\\ Agree \\-ment\end{tabular} & \begin{tabular}[c]{@{}r@{}r@{}} Near\\ Agree \\-ment \end{tabular} & Z-score & AUC \\ 
\midrule
URGENTIAPARSE & 0.228 & 0.790 & 0.800 & 0.802 & 0.900 & 0.894 & 0.900 & 0.928 & 2.514 & 0.879 \\
EMERGINET & 0.401 & 0.979 & 0.560 & 0.602 & 0.731 & 0.747 & 0.820 & 0.860 & 0.438 & 0.686 \\ 
TRIAGEMASTER & 0.637 & 1.180 & 0.370 & 0.005 & 0.618 & 0.613 & 0.554 & 0.696 & -3.511 & 0.642 \\ 
Real triage nurse & 1.393 & 1.834 & 0.080 & 0.024 & 0.303 & 0.275 & 0.303 & 0.498 & -4.343 & 0.776 \\
\bottomrule
\end{tabular}
\caption{Results of the main analysis aimed at comparing the triage processes according to a composite score}
\label{tab:table2}
\end{table}

\begin{figure}[!htbp]
\centering
    \includegraphics[width=\textwidth]{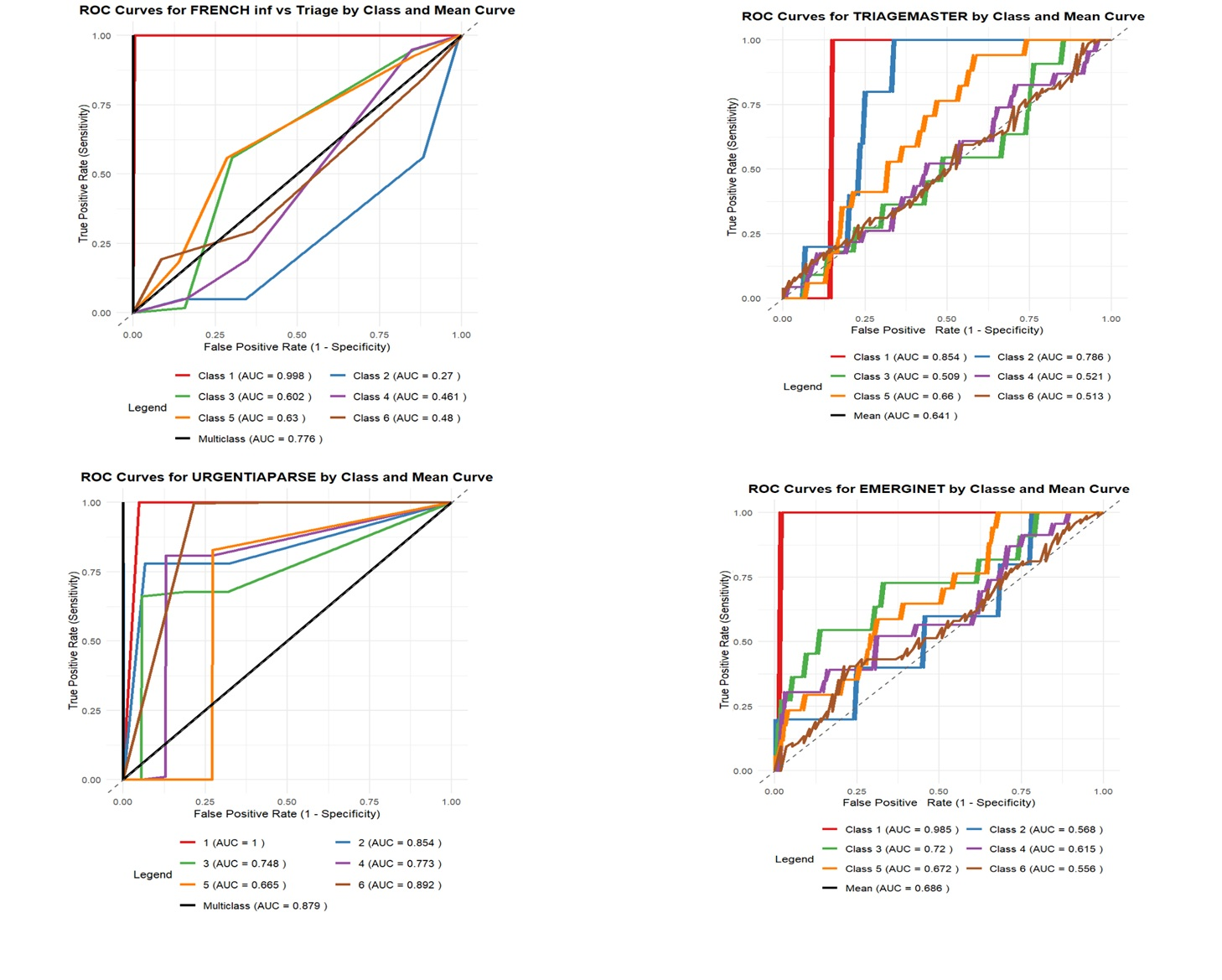}
    \caption{ROC curves for each triage process by Class and Mean Curve}
\label{fig:fig3}
\end{figure}

URGENTIAPARSE (LLM) led with a composite score of 2.514 (vs 4.902 for the FRENCH triage). EMERGINET (JEPA) followed with 0.438, then TRIAGEMASTER (NLP) with –3.511 (Table \ref{tab:table2}). Nurse triage scored lowest (–4.343). URGENTIAPARSE had strong F1\_Micro (0.899), F1\_Macro (0.894) and AUC (0.879). EMERGINET (scored 0.820, 0.747 and 0.686, respectively), TRIAGEMASTER and nurse triage lagged behind ( Fig. \ref{fig:fig3}).

\begin{figure}[!htbp]
\centering
    \includegraphics[width=\textwidth]{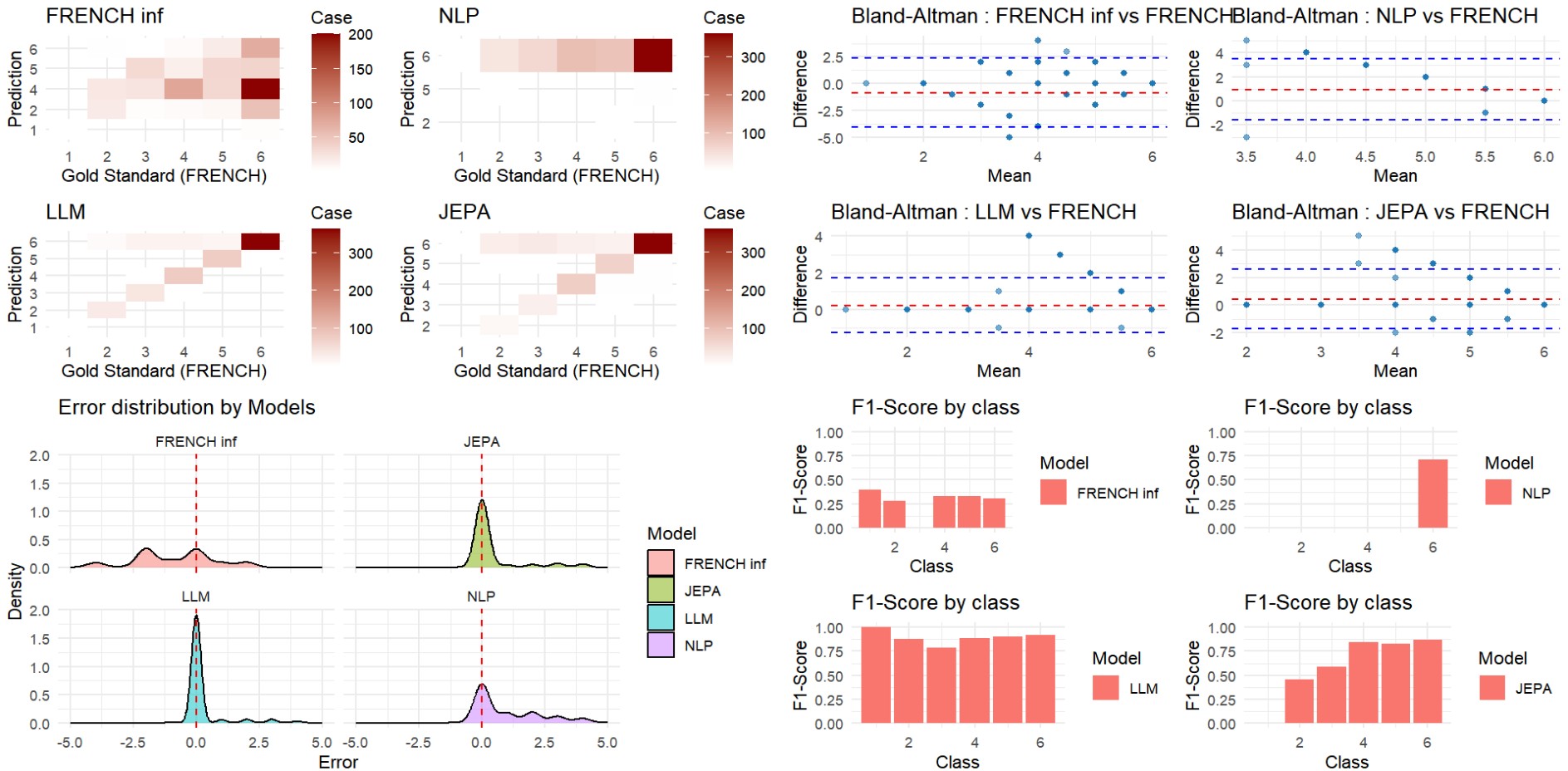}
    \caption{Primary analysis aimed at comparing the 3 AI models (TRIAGEMASTER, URGENTIAPARSE, EMERGINET) and nurse triage in their ability to predict FRENCH compared to ideal FRENCH}
\label{fig:fig4}
\end{figure}

Confusion matrices showed URGENTIAPARSE and EMERGINET slightly overestimated low triage levels, while TRIAGEMASTER underestimated and nurse triage was inconsistent. Bland-Altman and error analyses confirmed low, centered errors for URGENTIAPARSE and EMERGINET. Both also had consistently strong class-level F1 scores, making them the most reliable models (Fig. \ref{fig:fig4}).

\subsection{Secondary analyses}
\subsubsection{Calibration and reliability}
\begin{figure}[!htbp]
\centering
    \includegraphics[width=\textwidth]{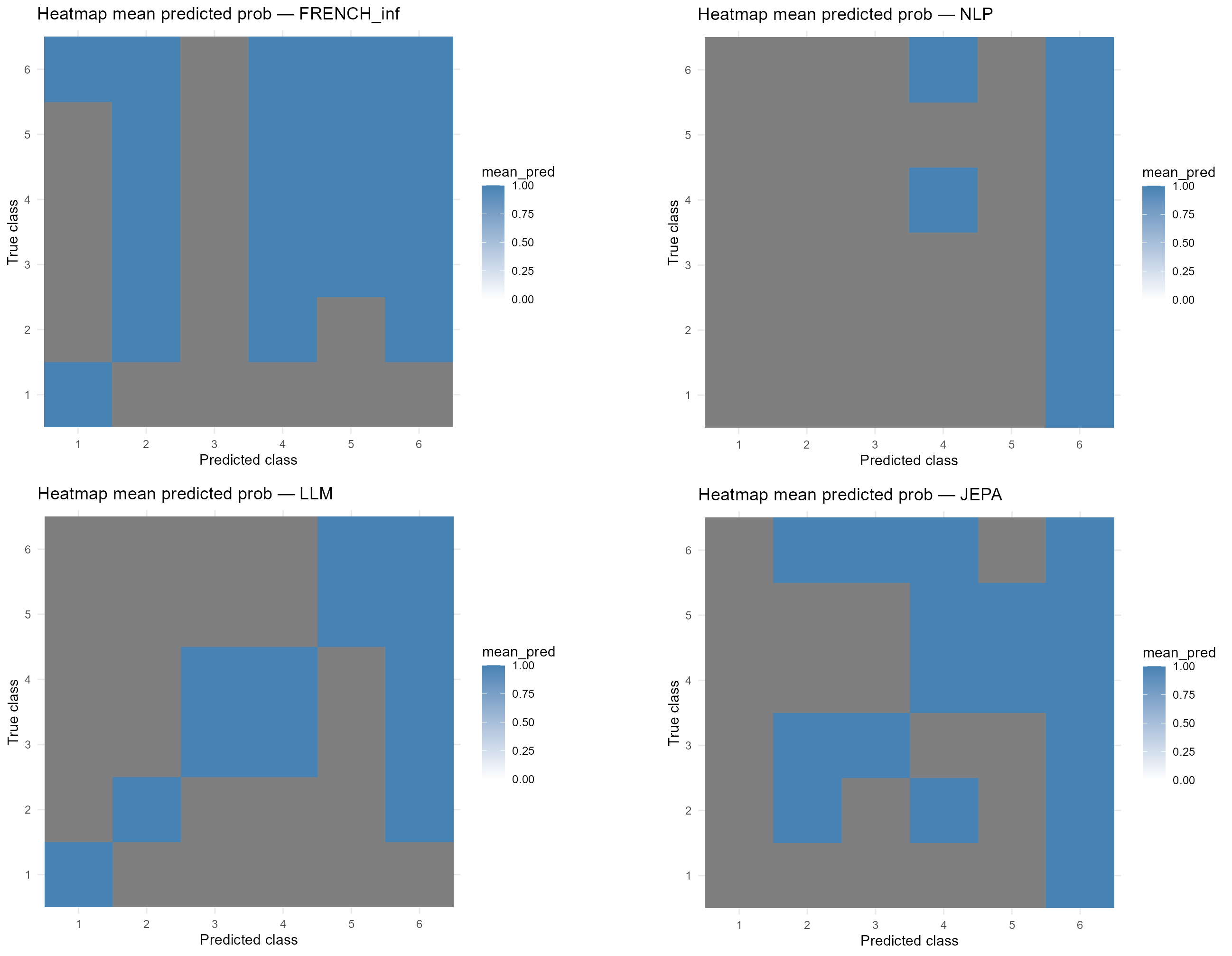}
    \caption{Class-Specific Calibration Heatmaps of Predicted Probabilities}
\label{fig:fig5}
\end{figure}

To evaluate predictive reliability, we assessed class-specific calibration using reliability diagrams represented as heatmaps (Figure \ref{fig:fig5}) and examined calibration tables stratified by prediction bins. 

Calibration analysis revealed heterogeneous reliability across classes. Class 1 predictions were overconfident, with sparse coverage of intermediate probabilities. Class 2 was dominated by high-probability predictions, indicating limited use of moderate confidence levels. Class 3 exhibited a partially calibrated, diagonal trend but showed inconsistencies across bins. Class 4 displayed the most balanced calibration, with broader coverage and closer agreement between predicted and observed frequencies. Together, these results highlight substantial class-specific differences in calibration performance.

\subsubsection{Risk distribution}
The nurse triage demonstrates relatively consistent distributional patterns across most triage scenarios, with distinct modal peaks for each class and moderate overlap between adjacent classes. The predicted probability distributions show clear separation, particularly in the middle range of the probability spectrum, suggesting reasonable discriminative capacity.\\
In contrast, TRIAGEMASTER (the NLP-based approach) exhibits more variable performance across different triage classes. Notable gaps appear in certain class predictions (particularly evident in panels 1 and 3), indicating potential limitations in the model's ability to consistently predict across all triage categories. The distributions show greater concentration around specific probability values, suggesting either higher confidence in certain predictions or potential overfitting to particular probability ranges.\\
URGENTIAPARSE (the LLM approach) displays markedly different characteristics compared to the other methodologies. Most notably, the LLM model shows substantial gaps in class representation across different panels, with several triage classes appearing absent in certain prediction scenarios. This sparse representation pattern suggests that the LLM may exhibit more selective or confident predictions, potentially indicating either superior discrimination between clearly distinct cases or limitations in handling ambiguous triage scenarios. The distributions that are present tend to be more concentrated and exhibit distinct modal peaks, suggesting high confidence when predictions are made. \\
EMERGINET (the JEPA model) displays heterogeneous patterns with some classes showing broad, diffuse distributions while others exhibit sharp, concentrated peaks, indicating variable predictive confidence across different triage classes.\\

\begin{figure}[!htbp]
\centering
    \includegraphics[width=\textwidth]{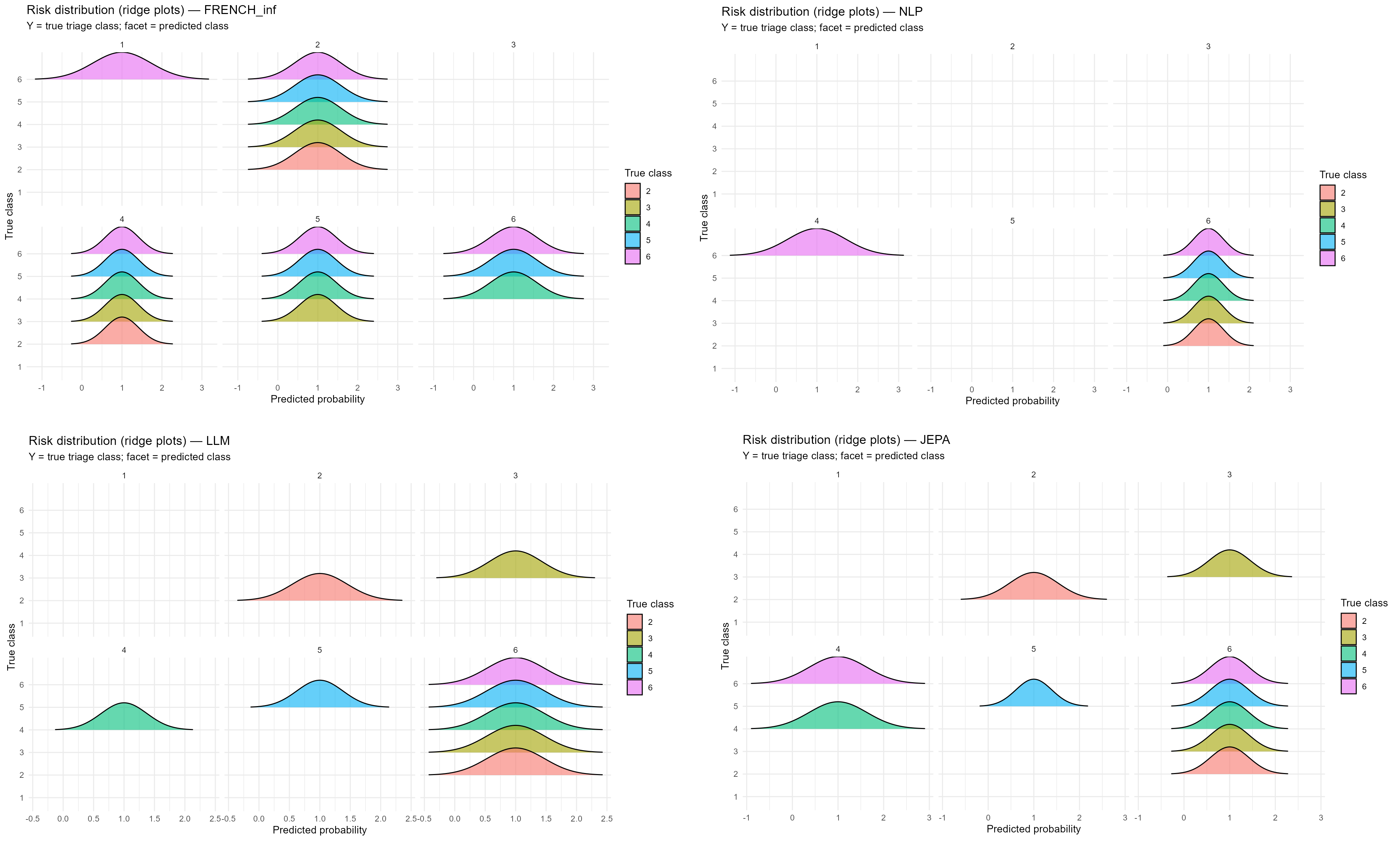}
    \caption{Ridge Plot Analysis of Classification Model Calibration Performance}
\label{fig:fig6}
\end{figure}

The distinct performance characteristics of the LLM approach, particularly its sparse prediction patterns, warrant careful consideration for clinical implementation. While this may indicate superior precision in clear-cut cases, the absence of predictions for certain class combinations could pose challenges in comprehensive triage coverage, requiring careful validation in clinical settings to ensure adequate coverage across all patient presentations (Fig.\ref{fig:fig6}).

\subsubsection{GEMSA Performance}
URGENTIAPARSE outperformed TRIAGEMASTER and EMERGINET in GEMSA prediction, with the best metrics (MAE 0.082, RMSE 0.402, kappa 0.863, Spearman 0.864). Its high F1 scores and strong agreement confirmed its accuracy and reliability, earning the top overall score of 2.382 (Table \ref{tab:table3}). \\

\begin{table}[!htbp]
\centering
\begin{tabular}{l *{9}{c}}
\toprule
Model & MAE & RMSE & Kappa & \begin{tabular}[c]{@{}r@{}} Spear\\ -man \end{tabular} & \begin{tabular}[c]{@{}r@{}} F1\\ Micro\end{tabular} & \begin{tabular}[c]{@{}r@{}} F1\\ Macro\end{tabular} & \begin{tabular}[c]{@{}r@{}r@{}} Exact\\ Agree \\-ment\end{tabular} & \begin{tabular}[c]{@{}r@{}r@{}} Near\\ Agree \\-ment \end{tabular} & Z-score \\
\midrule
URGENTIAPARSE    & 0.082 & 0.402 & 0.863 & 0.864 & 0.957 & 0.628 & 0.957 & 0.960 & 2.473 \\
TRIAGEMASTER    & 0.364 & 0.863 & 0.185 & 0.240 & 0.811 & 0.198 & 0.811 & 0.840 & -2.056 \\
EMERGINET   & 0.988 & 1.009 & 0    & 0    & 0.027 & 0.009 & 0.027 & 0.985 & -4.399 \\
\bottomrule
\end{tabular}
\caption{Results of the secondary analysis aimed at comparing the 3 AI models in their ability to predict GEMSA compared to the actual GEMSA according to a composite score}
\label{tab:table3}
\end{table}

\begin{figure}[!htbp]
\centering
    \includegraphics[width=\textwidth]{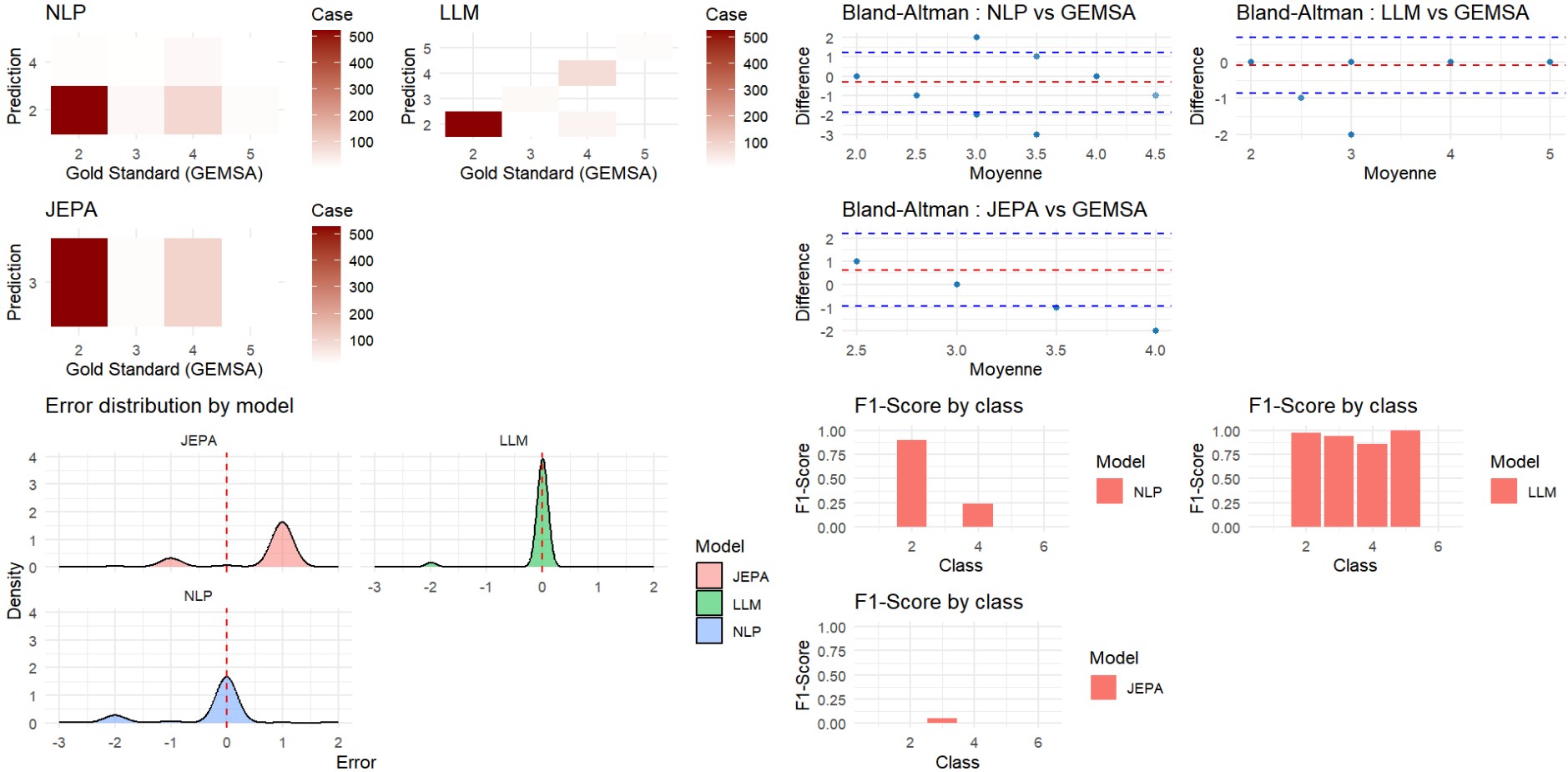}
    \caption{Secondary analysis aimed at comparing the 3 AI models (TRIAGEMASTER, URGENTIAPARSE, EMERGINET) in their ability to predict GEMSA compared to actual GEMSA}
\label{fig:fig7}
\end{figure}

Confusion matrices showed URGENTIAPARSE had the most accurate predictions, EMERGINET was moderately accurate, and TRIAGEMASTER's were scattered. Bland-Altman analysis confirmed strong agreement for URGENTIAPARSE, moderate for EMERGINET, and large deviations for TRIAGEMASTER (Fig. \ref{fig:fig7}).

Error analysis showed EMERGINET had small, focused errors; URGENTIAPARSE’s errors were slightly broader but centered; and TRIAGEMASTER had the widest and least centered errors. URGENTIAPARSE had consistently high F1 scores, EMERGINET was strong but less consistent, and TRIAGEMASTER was uneven.

\subsubsection{Structured data V.S. Unstructured data Performance}

The FRENCH model scored perfectly on all metrics and data types. URGENTIAPARSE closely followed with strong agreement on structured data but slightly lower on transcripts. EMERGINET showed moderate alignment on transcripts, while TRIAGEMASTER performed poorly on both data types. Overall, structured data led to better results, especially for URGENTIAPARSE (Table \ref{tab:table4}).

\begin{table}[!htbp]
\centering
\begin{tabular}{ll *{9}{c}}
\toprule
Model & Data & MAE & RMSE & Kappa & \begin{tabular}[c]{@{}r@{}} Spear\\ -man \end{tabular} & \begin{tabular}[c]{@{}r@{}} F1\\ Micro\end{tabular} & \begin{tabular}[c]{@{}r@{}} F1\\ Macro\end{tabular} & \begin{tabular}[c]{@{}r@{}r@{}} Exact\\ Agree \\-ment\end{tabular} & \begin{tabular}[c]{@{}r@{}r@{}} Near\\ Agree \\-ment \end{tabular} & Z-score \\
\midrule
URGENTIAPARSE  & Structured   & 0.213 & 0.748 & 0.821 & 0.826 & 0.903 & 0.898 & 0.903 & 0.928 & 1.806 \\
URGENTIAPARSE  & Unstructured & 0.240 & 0.807 & 0.790 & 0.790 & 0.893 & 0.891 & 0.893 & 0.924 & 1.628 \\
EMERGINET & Unstructured & 0.516 & 1.240 & 0.430 & 0.560 & 0.791 & 0.539 & 0.791 & 0.843 & -0.993 \\
EMERGINET & Structured   & 0.510 & 1.247 & 0.420 & 0.559 & 0.807 & 0.551 & 0.807 & 0.836 & -1.357 \\
TRIAGEMASTER  & Structured   & 0.916 & 1.557 & 0.057 & 0.160 & 0.572 & 0.1898 & 0.572 & 0.712 & -4.790 \\
TRIAGEMASTER  & Unstructured & 0.991 & 1.629 & -0.003 & 0.001 & 0.545 & 0.118 & 0.545 & 0.691 & -4.949 \\
\bottomrule
\end{tabular}
\caption{Results of a secondary analysis aimed at comparing 4 sorting processes according to a composite score}
\label{tab:table4}
\end{table}

\section{Discussion}
From June to December 2024, 657 nurse interviews were analyzed from over 73,000 ED visits. URGENTIAPARSE was the top-performing AI model, outperforming EMERGINET, TRIAGEMASTER, and nurse triage. It showed high accuracy and strong alignment with FRENCH and GEMSA standards, especially when using structured data—emphasizing the importance of input format.
\subsection{Limitations}
The study was limited by a small sample (0.9\% of patients), few participating nurses, and potential bias from its university hospital setting. Data reliability was affected by variable recording quality and equipment loss. \\
URGENTIAPARSE was the top performer using the FRENCH triage scale but showed signs of overfitting. TRIAGEMASTER had stable, well-generalized results. External models like Mistral 7B and Llama 3 8B performed moderately, while larger models like Mixtral 8x7B underperformed\cite{pmlr-v259-guerra-adames25a}. In another study, Clinical BELGPT-2 led with an F1-score of 0.62, while models like XGBoost/TF-IDF and LightGBM/TF-IDF had more modest scores\cite{Avalos2024DetectingHB}. EMERGINET also showed overfitting. Improving URGENTIAPARSE’s generalization may require regularization, data augmentation, or model simplification, and addressing hallucinations could involve fine-tuning or prompt refinement \cite{Farquhar2024DetectingHI}, TimeLLM approaches \cite{Jin2023TimeLLMTS}, or backbone language models reprogrammation. \\
Another limitation is the exclusion of repeat patients, which hinders analysis of long-term care and triage bias. Hospital-specific protocols also limit generalizability. Once deployed, the models will run automatically without user input. \\
Additionally, few independent studies have evaluated the FRENCH triage scale beyond its original validations\cite{Taboulet2009TriageWT}. In 2009, its predictive validity for hospitalization showed an AUC of 0.858, and significant correlations with care consumption and waiting time. A 2019 validation with 29,423 patients found similar results (AUC of 0.83), with moderate but significant correlations between triage level and resource utilization or testing\cite{Taboulet2019ValiditDL}. Thus, even with the metrics of the FRENCH validation studies, the URGENTIAPARSE model remains with superior performances (e.g. AUC) than nurses that have executed the triage during those studies.
\subsection{Conclusion}
This study compared three AI models using real nurse-patient data and a national triage standard, finding URGENTIAPARSE as the most accurate. It highlights the need for broader validation, real-time testing, and adaptive learning, supporting LLM-based AI to enhance triage safety and consistency.

\bibliographystyle{unsrt}  
\bibliography{references}

\end{document}